\documentclass{article}

\usepackage[final]{neurips_2019}

\usepackage[utf8]{inputenc} 
\usepackage[T1]{fontenc}    
\usepackage{hyperref}       
\usepackage{url}            
\usepackage{booktabs}       
\usepackage{amsfonts}       
\usepackage{nicefrac}       
\usepackage{microtype}      
\usepackage{comment}
\usepackage{chngpage}

\author{
Hassan Kané \\
  WL Research \\
  \texttt{hassanmohamed@alum.mit.edu} \\
  \And
  Yusuf Kocyigit \\
  WL Research \\
  Boğaziçi University \\
  \texttt{ yusuf.kocyigit@boun.edu.tr} \\
   \And
  Ali Abdalla \\
  WL Research\\
  \texttt{aabdalla@alum.mit.edu} \\
   \And
  Pelkins Ajanoh \\
   WL Research \\
  \texttt{pelkins@alum.mit.edu} \\
  \And
  Mohamed Coulibali \\
  WL Research\\
  Laval University \\
  \texttt{mohamed-konoufo.coulibaly.1@ulaval.ca} \\
}
\title{Towards Neural Similarity Evaluators}

\begin{document}

\maketitle

\begin{abstract}

We review the limitations of BLEU and ROUGE -- the most popular metrics used to assess reference summaries against hypothesis summaries, and come up with criteria for what a good metric should behave like and propose concrete ways to use recent Transformers-based Language Models to assess reference summaries against hypothesis summaries.
    
\end{abstract}


\section{Introduction}
Evaluation metrics play a central role in the machine learning community. They direct the efforts of the research community and are used to define the state of the art models. In machine translation and summarization, the two most common metrics used for evaluating similarity between candidate and reference texts are BLEU \citep{papineni2002bleu} and ROUGE \citep{lin2004rouge}. Both approaches rely on counting the matching n-grams in the candidates summary to n-grams in the reference text. BLEU is precision focused while ROUGE is recall focused. 

These metrics have posed serious limitations and have already been criticized by the academic community [to include citations]. In this work, we formulate an empirical criticism of BLEU and ROUGE, establish a criteria that a sound evaluation metric should have and propose concrete ways to use recent advances in NLP to design data-driven metric addressing the weaknesses found in BLEU and ROUGE and scoring high on the criteria for a sound evaluation metric.

\section{Related Work}

\subsection{BLEU, ROUGE and n-gram matching approaches}

BLEU (Bilingual Evaluation Understudy) \citep{papineni2002bleu} and ROUGE (Recall-Oriented Understudy for Gisting Evaluation) \citep{lin2004rouge} have been used to evaluate many NLP tasks for almost two decades. The general acceptance of these methods depend on many factors including their simplicity and the intuitive interpretability. Yet the main factor is the claim that they highly correlate with human judgement \citep{papineni2002bleu}. This has been criticised extensively by the literature and the shortcomings of these methods have been widely studied. Reiter \citep{reiter2018structured} , in his structured review of BLEU, finds a low correlation between BLEU and human judgment. Callison et al \citep{callison2006re} examines BLEU in the context of machine translation  and find that BLEU does neither correlate with human judgment on adequacy(whether the hypothesis sentence adequately captures the meaning of the reference sentence) nor fluency(the quality of language in a sentence). Sulem et al \citep{sulem2018bleu} examines BLEU in the context of text simplification on grammaticality, meaning preservation and simplicity and report BLEU has very low or in some cases negative correlation with human judgment.

\subsection{Transformers, BERT and GPT}

Language modeling has become an important NLP technique thanks to the ability to apply it to various NLP tasks as explained in Radford et al \citep{radford2019language}. There are two leading architectures for language modeling Recurrent Neural Networks (RNNs)\citep{mikolov2010recurrent} and Transformers \citep{vaswani2017attention} . RNNs handle the input tokens, words or characters, one by one through time to learn the relationship between them, whereas, transformers receive a segment of tokens and learn the dependencies between them using an attention mechanism. 

\subsection{Model-based metrics}

While BLEU and ROUGE are defined in a discrete space new evaluation metric can be defined in this continuous space. BERTscore \citep{zhang2019bertscore} uses word embeddings and cosine similarity to create a score array and use greedy matching to maximize the similarity score. Sentence Mover’s Similarity \citep{clark2019sentence} uses the mover similarity, Wasserstein distance, between sentence embedding generated from averaging the word embeddings in a sentence. 

One other evaluation method proposed is RUSE \citep{shimanaka2018ruse} this method proposes embedding both sentences separately and pooling them to a given size. After that they use a pre trained MLP to predict on different tasks. This quality estimator metric is then proposed to be used in language evaluation. 

Our proposed methodology is to take neural language evaluation beyond architecture specifications. We are proposing a framework in which an evaluator's success can be determined.

\section{Challenges with BLEU and ROUGE}
In this part, we discuss three significant limitations of BLEU and ROUGE. These metrics can assign: High scores to semantically opposite translations/summaries, Low scores to semantically related translations/summaries and High scores to unintelligible translations/summaries.

\subsection{High score, opposite meanings}

Suppose that we have a reference summary s1. By adding a few negation terms to s1, one can create a summary s2 which is semantically opposite to s1 but yet has a high BLEU/ROUGE score. 

\subsection{Low score, similar meanings}

In addition not to be sensitive to negation, BLEU and ROUGE score can give low scores to sentences with equivalent meaning. If s2 is a paraphrase of s1, the meaning will be the same ;however, the overlap between words in s1 and s2 will not necessarily be significant. 

\subsection{High score, unintelligible sentences}

A third weakness of BLEU and ROUGE is that in their simplest implementations, they are insensitive to word permutation and can give very high scores to unintelligible sentences.

\section{Assessing evaluation metrics}

\subsection{Metric Scorecard}

To overcome the previously highlighted challenges and provide a framework by which metrics comparing reference summaries/translation can be assessed and improved, we established first-principles criteria on what a good evaluator should do. 

The first one is that it should be highly correlated with human judgement of similarity. The second one is that it should be able to distinguish sentences which are in logical contradiction, logically unrelated or in logical agreement. The third one is that given s1, s2 which are semantically similar, eval(s1,s2) > eval(s1,s2(corrupted) > eval(s1,s2(more corrupted)) where corruption here includes removing words or including grammatical mistakes.

\subsection{Implementing metrics satisfying scorecard}
\subsubsection{Semantic Similarity}
Starting from the RoBERTa large pre-trained model \citep{liu2019roberta} , we finetune it to predict sentence similarity (0-5 scale) on the STS-B benchmark dataset (8628 sentence pairs). 

\subsubsection{Logical Equivalence}
For logical inference, we start with a pretrained RoBERTa \citep{liu2019roberta} model and finetune it using the Multi-Genre Natural Language Inference Corpus (433k sentence pairs) \citep{williams2017broad}. The accuracy of the pre-trained model on the development set is \textbf{0.9060}.

\subsection{Experiments}
After highlighting challenges with BLEU and ROUGE, presenting alternative metrics for each of our criteria, we now proceed to score them against the previously mentioned scorecard.

\subsubsection{Semantic similarity experiments} 
We assessed how well BLEU and ROUGE correlated with human judgement of similarity between pairs of paraphrased sentences and compared their performance to a RoBERTa model finetuned for semantic similarity (Table 1).

\begin{table}
\centering 
    \caption{Correlation with human judgement of similarity on STS-B Benchmark development set}
	\begin{tabular}{|l|l|l|l|l|l|l|}
		\hline
                  & ROUGE & BLEU & RoBERTa-STS  \\ \hline
STS-B &          0.47  &         0.31  &         \textbf{0.92}     \\ \hline
	\end{tabular}
    \label{tab:correlations}
\end{table}

\subsubsection{Logical Entailment experiments}

For 300 sentences from MNLI, we assessed for each sentence s1, how well BLEU, ROUGE and RoBERTa trained on STS-B would rank sentences in contradiction, neutral relation or entailment (Table 2).

\begin{table}
\centering 
    \caption{Results of grammatical error experiments}
	\begin{tabular}{|l|l|l|l|l|l|l|}
		\hline
                  & ROUGE & BLEU & RoBERTa-STS  \\ \hline
Spearman's RC &          0.528  &         0.472  &         \textbf{0.718}     \\ \hline
Kendall's Tau &          0.478  &         0.419  &         \textbf{0.667}     \\ \hline
	\end{tabular}
    \label{tab:kendal_tau}
\end{table}

\subsubsection{Robustness to grammatical errors experiments}

For assessing the third criteria. We start with 3479 sentence pairs which both include paraphrases and regular similar sentences. We introduce random corruptions such as random insertion, deletion and grammatical errors as in \citep{zhao2019improving}. We report results on table 3.

\begin{table}
\centering 
    \caption{Results of logical entailment experiments}
	\begin{tabular}{|l|l|l|l|l|l|l|}
		\hline
                  & ROUGE & BLEU & RoBERTa-STS  \\ \hline
Spearman's RC &          0.255  &         0.216  &         \textbf{0.744}     \\ \hline
Kendall's Tau &          0.215  &         0.186  &         \textbf{0.69}     \\ \hline
	\end{tabular}
    \label{tab:kendal_tau}
\end{table}

\section{Conclusion}

In this work, we have established a framework to assess metrics comparing the quality of reference and hypothesis summary/translations. Based on these criteria, we compare evaluators using recent Transformers advance to BLEU and ROUGE and highlight their potential replace BLEU and ROUGE.

\bibliographystyle{abbrvnat}
\bibliography{main.bib}

\end{document}